\documentclass[letterpaper, 10 pt, conference]{ieeeconf}  

\usepackage{graphicx}
\usepackage{multirow}
\usepackage[table,xcdraw]{xcolor}
\usepackage{float}
\usepackage{booktabs}
\makeatletter

\newcommand{\Rmnum}[1]{\expandafter\@slowromancap\romannumeral #1@}
\makeatother
\usepackage{cite}
\usepackage{url}
\usepackage{amsmath}
\usepackage{amssymb}
\usepackage{hyperref}

\IEEEoverridecommandlockouts    

\title{\LARGE \bf
Towards Cross-Subject EMG Pattern Recognition via Dual-Branch Adversarial Feature Disentanglement
}

\author{Xinyue Niu$^{1}$and Akira Furui$^{1}$
\thanks{This work was supported by JST BOOST, Japan Grant Number JPMJBS2424.}
\thanks{$^{1}$Xinyue Niu and Akira Furui are with the Graduate School of Advanced Science and Engineering, Hiroshima University, Higashi-hiroshima 739-8521, Japan (e-mail:
   {\tt\small xinyueniu@hiroshima-u.ac.jp; akirafurui@hiroshima-u.ac.jp}).}%
}

\begin{document}

\maketitle
\thispagestyle{empty}
\pagestyle{empty}

\begin{abstract}

Cross-subject electromyography (EMG) pattern recognition faces significant challenges due to inter-subject variability in muscle anatomy, electrode placement, and signal characteristics.
Traditional methods rely on subject-specific calibration data to adapt models to new users, an approach that is both time-consuming and impractical for large-scale, real-world deployment.
This paper presents an approach to eliminate calibration requirements through feature disentanglement, enabling effective cross-subject generalization. 
We propose an end-to-end dual-branch adversarial neural network that simultaneously performs pattern recognition and individual identification by disentangling EMG features into pattern-specific and subject-specific components.
The pattern-specific components facilitate robust pattern recognition for new users without model calibration, while the subject-specific components enable downstream applications such as task-invariant biometric identification.
Experimental results demonstrate that the proposed model achieves robust performance on data from unseen users, outperforming various baseline methods in cross-subject scenarios.
Overall, this study offers a new perspective for cross-subject EMG pattern recognition without model calibration and highlights the proposed model's potential for broader applications, such as task-independent biometric systems.

\end{abstract}

\section{Introduction}

Electromyography (EMG)-based pattern recognition has been widely applied across various fields, including prosthetic control~\cite{resnik2018evaluation}, human-computer interaction~\cite{xiong2021deep}, rehabilitation~\cite{cesqui2013emg}.
A critical challenge in this technology is achieving robust and accurate recognition across different individuals.
Due to physiological variability such as muscle anatomy, subcutaneous fat thickness, and skeletal structure, EMG characteristics exhibit significant inter-subject discrepancies, which limit the generalization capability of conventional methods based on subject-specific data~\cite{karnam2022emghandnet,khushaba2021long}.

To address this issue, most existing approaches rely on model calibration, which requires substantial labeled data from each new individual to fine-tune the model~\cite{cote2019deep,zou2021transfer}.
While this strategy can improve accuracy, it imposes significant burdens on both users and researchers, as the process of collecting and labeling data is time-consuming and impractical.
Moreover, this reliance on calibration data undermines the scalability of these methods in real-world applications that requires seamless deployment across diverse users.

Domain generalization (DG) is a machine learning paradigm designed to train models that generalize effectively across multiple domains without requiring access to unseen domains during the training phase~\cite{wang2022generalizing}. In the context of EMG pattern recognition, each subject can be considered as a distinct domain, and DG aims to learn subject-invariant features that are transferable across different individuals. This approach offers a promising solution to the limitations of conventional calibration-based methods, as it eliminates the need for collecting additional data from new users.

However, research applying DG to EMG pattern recognition remains limited.
Li \textit{et al.}~\cite{10025816} proposed a unified framework combining DG and unsupervised domain adaptation for cross-subject EMG recognition, where DG is used to extract user-generic features enabling preliminarily adaptation to unseen users, though this approach still requires data from these users for optimal results.  
Ye \textit{et al.}~\cite{ye2023cross} introduced a DG method using a dynamic network to extract more robust features, successfully eliminating the need for unseen user data but incorporating more complex network modules.
More recently, a notable contribution by Fan \textit{et al.}~\cite{fan2024surface} introduced the innovative concept of EMG feature disentanglement, demonstrating significant improvements in cross-subject EMG pattern recognition performance and showing potential for various downstream tasks.
Despite these promising outcomes, their study has two primary limitations.
First, their research relies on high-density EMG, which is costly and less practical compared to the sparse multi-channel systems commonly used in real-world applications.
Second, their approach is constrained by its reliance on handcrafted features and additional classifiers, potentially limiting its adaptability and scalability compared to fully automated end-to-end methods.

In this paper, we extend EMG feature disentanglement to multi-channel EMG signals, which is adopted in real-world scenarios due to their practicality, lower cost, and ease of implementation.
Inspired by~\cite{han2020disentangled} and~\cite{liu2018multi}, we propose an end-to-end dual-branch adversarial neural network that integrates feature extraction, feature disentanglement, and pattern recognition into a unified framework.
The key innovation lies in our approach to disentangling EMG features into pattern-specific and subject-specific components through adversarial learning, enabling effective isolation of domain-invariant features.
Unlike previous methods, our end-to-end architecture automatically learns optimal feature representations directly from raw EMG signals, eliminating the need for handcrafted features and model calibration. 
Through comprehensive experiments using 40 subjects, we demonstrate that this approach achieves robust cross-subject EMG pattern recognition while maintaining practical applicability for real-world deployment.

\section{Proposed Method}

\subsection{Problem Settings}

Let $\mathcal{D} = \{(\mathbf{X}_i, \mathbf{y}_i^\mathrm{p}, \mathbf{y}_i^\mathrm{s})\}_{i=1}^N$ be a training dataset containing $N$ samples from multiple subjects performing different gesture patterns. 
Each sample $\mathbf{X}_i \in \mathbb{R}^{L \times D}$ represents a multi-channel EMG signal segment, where $L$ is the window length and $D$ is the number of channels. 
For each sample, $\mathbf{y}_i^\mathrm{p} \in \{0,1\}^{N_\mathrm{p}}$ and $\mathbf{y}_i^\mathrm{s} \in \{0,1\}^{N_\mathrm{s}}$ are one-hot encoded labels for $N_\mathrm{p}$ gesture patterns and $N_\mathrm{s}$ subjects, respectively. 
Our goal is to learn a model that generalizes well to EMG signals from unseen subjects without requiring additional calibration data.

\subsection{Overall Framework}

\begin{figure}[t] 
    \centering 
    \includegraphics[width=0.48\textwidth]{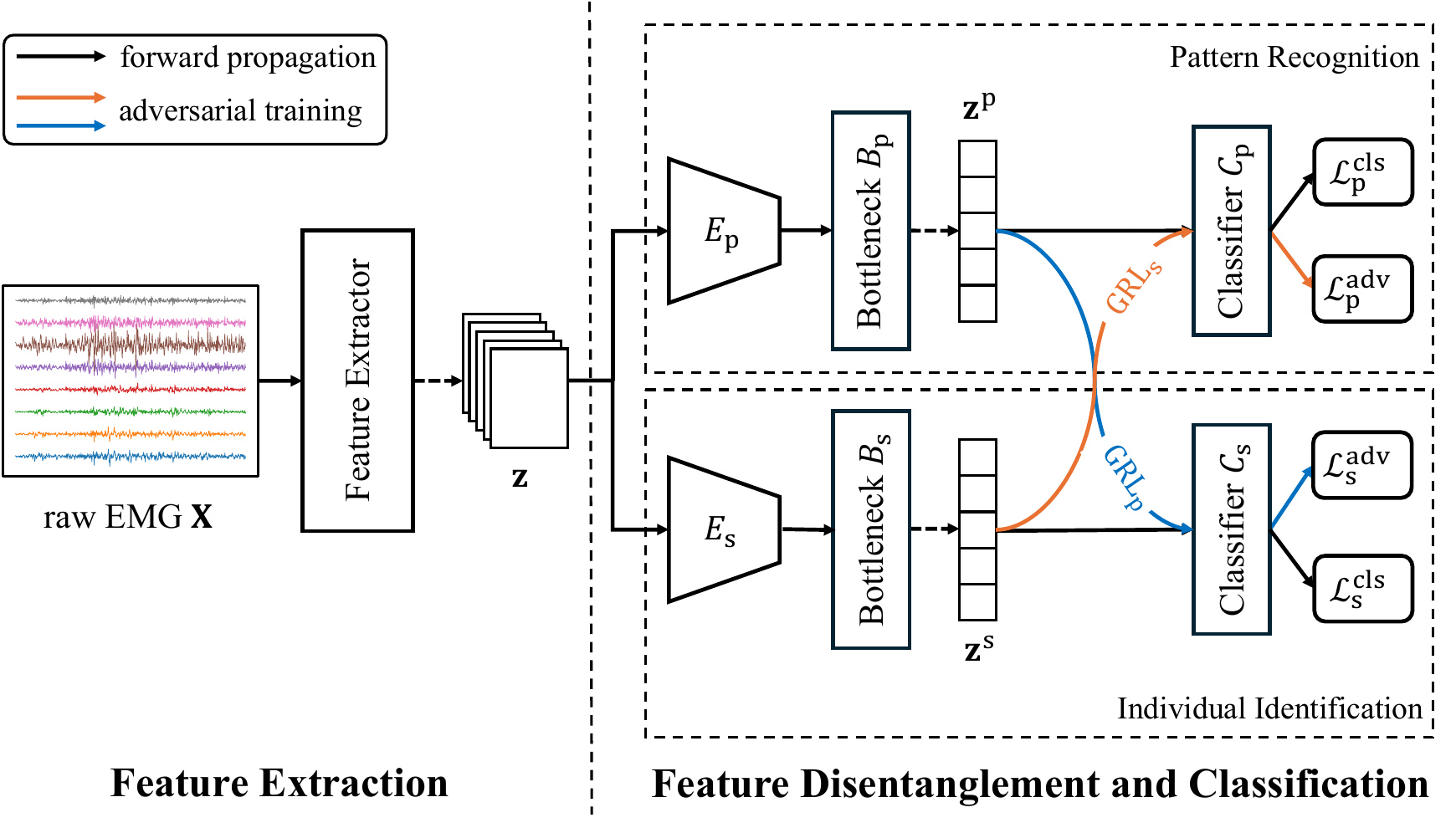} 
    \caption{Architecture of the proposed dual-branch adversarial neural network. The network consists of a shared feature extractor followed by two parallel branches for EMG pattern recognition and individual identification. Each branch incorporates adversarial training through gradient reversal layers (GRL) to disentangle pattern-specific and subject-specific features.} 
    \label{fig:method} 
\end{figure}

The overall structure of the proposed dual-branch adver-
sarial neural network is depicted in Fig.~\ref{fig:method}.
The network comprises three primary components: (1) a shared feature extractor that captures representations containing both pattern- and subject-related information (referred to as ``original features''), (2) a pattern recognition branch that disentangles pattern-specific features from the original features, and (3) an individual identification branch that disentangles subject-specific features.  

This architecture leverages multi-task learning to simultaneously address pattern recognition and individual identification while reducing the number of parameters through a shared feature extractor. This shared structure promotes the extraction of cleaner original features due to the regularizing effect of multiple learning objectives. The network incorporates adversarial training to ensure the disentangled features exhibit both cross-domain transferability (across subjects or patterns) and sufficient discriminative power for their respective tasks. This combination of multi-task learning and adversarial training enhances the robustness and generalizability of the proposed model.

\subsection{Original Feature Extraction}

The feature extractor transforms each input signal into high-level features $\mathbf{z}_i \in \mathbb{R}^{W \times C'}$, where $W$ and $C'$ represent the temporal width and number of feature channels.
This transformation is defined as:
\begin{equation}
\mathbf{z}_i=\Phi(\mathbf{X}_i).
\end{equation}
The feature extractor $\Phi(\cdot)$ is implemented using a series of convolutional layers, each followed by batch normalization (BN) and rectified linear unit (ReLU) activation.
The extracted features $\mathbf{z}_i$ encode both pattern- and subject-related information for subsequent disentanglement.

\subsection{Feature Disentanglement and Classification}

The extracted features $\mathbf{z}_i$ are then processed through two parallel mirror-designed branches: 
\begin{itemize}
\item \textbf{Pattern recognition branch} that extracts pattern-specific features and classifies them into predefined gesture categories
\item \textbf{Individual identification branch} that extracts subject-specific features and identifies individuals through predefined subject IDs.
\end{itemize}
Each branch consists of an encoder $E$, a bottleneck layer $B$, and a classifier $C$, with subscripts $\mathrm{p}$ and $\mathrm{s}$ denoting pattern recognition and subject identification components, respectively.
The encoders $E_\mathrm{p}$ and $E_\mathrm{s}$ are implemented using a series of convolutional blocks to extract pattern-specific and subject-specific representations from the original features. 
The bottleneck layers $B_\mathrm{p}$ and $B_\mathrm{s}$ are employed as linear layers to reduce feature dimensions, thereby lowering computational costs and parameter count while preserving essential information. In this study, we consider the outputs after the bottleneck layers as pattern-specific features $\mathbf{z}_i^\mathrm{p}$ and subject-specifc features $\mathbf{z}_i^\mathrm{s}$, which can be represented as:
\begin{equation}
\mathbf{z}_i^\mathrm{p} = B_\mathrm{p}(E_\mathrm{p}(\mathbf{z}_i)),
\end{equation}
\begin{equation}
\mathbf{z}_i^\mathrm{s} = B_\mathrm{s}(E_\mathrm{s}(\mathbf{z}_i)).
\end{equation}
The classifiers $C_\mathrm{p}$ and $C_\mathrm{s}$ are implemented as linear layers followed by softmax activation to perform the final classification. The network outputs probability distributions $\hat{\mathbf{y}}_i^\mathrm{p} \in [0,1]^{N_\mathrm{p}}$ and $\hat{\mathbf{y}}_i^\mathrm{s} \in [0,1]^{N_\mathrm{s}}$ over pattern classes and subjects, respectively. The classification process of each branch can be expressed as:
\begin{equation}
 \hat{\mathbf{y}}_i^\mathrm{p} = C_\mathrm{p}(\mathbf{z}_i^\mathrm{p}),
\end{equation}
\begin{equation}
\hat{\mathbf{y}}_i^\mathrm{s} = C_\mathrm{s}(\mathbf{z}_i^\mathrm{s}).
\end{equation}

To achieve effective disentanglement of pattern-specific and subject-specific features, we incorporate gradient reversal layers (GRL) into the network. 
A GRL acts as an identity function during forward propagation but reverses the gradient by multiplying it by a negative scalar during backpropagation. 
In the proposed method, two GRLs are introduced: $\text{GRL}_\mathrm{p}$ between $B_\mathrm{p}$ and $C_\mathrm{s}$, and $\text{GRL}_\mathrm{s}$ between $B_\mathrm{s}$ and $C_\mathrm{p}$. The adversarial predictions are given by:
\begin{equation}
\hat{\mathbf{y}}_i^\mathrm{adv,s} = C_\mathrm{s}(\text{GRL}_\mathrm{p}(\mathbf{z}_i^\mathrm{p})) \in [0,1]^{N_\mathrm{s}},
\end{equation}
\begin{equation}
\hat{\mathbf{y}}_i^\mathrm{adv,p} = C_\mathrm{p}(\text{GRL}_\mathrm{s}(\mathbf{z}_i^\mathrm{s})) \in [0,1]^{N_\mathrm{p}}.
\end{equation}
These adversarial branches minimize mutual information between pattern-specific and subject-specific feature spaces, forcing each branch to learn representations that are invariant to the other task's attributes.

\subsection{Loss Function}

The network is trained to minimize classification losses while maximizing adversarial losses to achieve effective feature extraction, disentanglement, and classification. The classification losses  $\mathcal{L}_{\mathrm{p}}^\mathrm{cls}$ and $\mathcal{L}_{\mathrm{s}}^\mathrm{cls}$ for pattern recognition and individual identification are implemented using cross-entropy as follows:
\begin{equation}
    \mathcal{L}_{\mathrm{p}}^\mathrm{cls}=-\sum_{i=1}^N\mathbf{y}_i^\mathrm{p}\log(\hat{\mathbf{y}}_i^\mathrm{p}),
\end{equation}
\begin{equation}
    \mathcal{L}_{\mathrm{s}}^\mathrm{cls}=-\sum_{i=1}^N\mathbf{y}_i^\mathrm{s}\log(\hat{\mathbf{y}}_i^\mathrm{s}).
\end{equation}

The $\mathcal{L}_\mathrm{s}^\mathrm{adv}$ and $\mathcal{L}_\mathrm{p}^\mathrm{adv}$ are the adversarial losses for the pattern recognition and individual identification branches, respectively. 
They are implemented using cross-entropy, with their gradient directions reversed by the GRL during backpropagation to minimize mutual information between the pattern-specific and subject-specific feature spaces:
\begin{equation}
    \mathcal{L}_\mathrm{s}^\mathrm{adv}=-\sum_{i=1}^N{\mathbf{y}}_i^\mathrm{s}\log(\hat{\mathbf{y}}_i^\mathrm{adv,s}),
\end{equation}
\begin{equation}
    \mathcal{L}_\mathrm{p}^\mathrm{adv}=-\sum_{i=1}^N{\mathbf{y}}_i^\mathrm{p}\log(\hat{\mathbf{y}}_i^\mathrm{adv,p}).
\end{equation}
The total training objective is formulated as:
\begin{equation}
    \mathcal{L}=\mathcal{L}_{\mathrm{p}}^\mathrm{cls}+\mathcal{L}_{\mathrm{s}}^\mathrm{cls}-\lambda_{\mathrm{p}}\mathcal{L}_\mathrm{p}^\mathrm{adv}-
    \lambda_{\mathrm{s}}\mathcal{L}_\mathrm{s}^\mathrm{adv},
    \label{eq:total loss}
\end{equation}
where $\lambda_{\mathrm{s}}$ and $\lambda_{\mathrm{p}}$ are the GRL layer hyperparameters used to balance the contributions of the adversarial losses. These values can be adjusted based on validation performance to ensure effective training.

\subsection{Training Strategy}

The shared feature extractor is updated jointly based on both classification and adversarial losses. For each epoch, the training process can be summarized as two step as follows:

\begin{enumerate}
    \item Update the pattern recognition branch and its associated adversarial branch based on $\mathcal{L}_{\mathrm{p}}^\mathrm{cls}$ and $\mathcal{L}_\mathrm{p}^\mathrm{adv}$.
    \item Update the individual identification branch and its adversarial branch based on $\mathcal{L}_{\mathrm{s}}^\mathrm{cls}$ and $\mathcal{L}_\mathrm{s}^\mathrm{adv}$.
\end{enumerate}

The number of iterations for each step can be adjusted based on validation performance. 
In this study, we perform one iteration per step. 
This alternating optimization process continues until the model converges. 
After convergence, the pattern recognition branch (without adversarial branch) can be directly applied to new users without requiring additional calibration.

\section{Experiments}

Experiments were conducted to evaluate the effectiveness of the proposed method on cross-subject EMG pattern recognition by comparing its performance with baseline methods under various conditions.

\subsection{Dataset}
The Gesture Recognition and Biometrics ElectroMyogram (GRABMyo) dataset is an open-access resource comprising data from 43 subjects collected over three days while performing 17 hand and finger gestures. The EMG signals were recorded from 8 forearm channels and 6 wrist channels at 2048 Hz. For each gesture, seven trials were recorded, each lasting 5 seconds.

In this study, we utilized forearm data collected on the first day. 
We selected 40 subjects for our experiments, excluding three subjects who showed consistently poor performance (accuracy below 50\%) in preliminary cross-validation tests.
The selected subject IDs were 1–23, 25–40, and 41. 
For gesture recognition tasks, we randomly selected six gestures: four finger gestures (gesture IDs: 2, 3, 4, 5) and two hand gestures (gesture IDs: 16, 17).
The raw EMG signals were preprocessed using a sliding window approach with a window length of 200 ms and a step size of 10 ms, resulting in input samples with dimensions of $408 \times 8$.

We evaluated the model performance using 4-fold cross-validation, where the 40 subjects were randomly divided into four groups of 10 subjects each. For each fold, data from 30 subjects were used for training, and the remaining 10 subjects' data were used for testing. This validation scheme with a relatively large test group enables us to better assess the model's generalization capability across a diverse range of subjects, compared to previous studies that typically employed leave-one-out cross-validation or smaller test groups.

\subsection{Implementation Details}

The network architecture consists of a shared feature extractor followed by two parallel branches. The feature extractor transforms raw input samples into original features through a convolutional block.
This block comprises a convolutional layer (kernel size: 5, input channels: 8, output channels: 32), followed by batch normalization (BN), ReLU activation, and max-pooling (pool size: 2, stride: 2), producing features of size $204 \times 32$.

The two parallel branches are designed with mirror-symmetric structures. Taking the pattern recognition branch as an example, it consists of an encoder, a bottleneck layer, and a classifier. The encoder contains two convolutional blocks. The first block has a convolutional layer (kernel size: 3, input channels: 32, output channels: 64), followed by BN, ReLU activation, and max-pooling (pool size: 2, stride: 2). The second block follows a similar structure with different parameters (kernel size: 5, input channels: 64, output channels: 128). The bottleneck layer comprises a linear layer with 256 units, followed by BN, ReLU activation, and dropout (rate: 0.5). The classifier consists of a fully connected layer (output units: 6 patterns or 30 subjects, depending on the task) with softmax activation.

The hyperparameters of the GRL layers are implemented as functions of the training progress, denoted as $\lambda_{\mathrm{s}}(p)$ and $\lambda_{\mathrm{p}}(p)$, which are linearly increased according to:
\begin{equation}
    \lambda_{\mathrm{s}}(p) = \lambda_{\mathrm{s}}^\mathrm{init} + p(\lambda_{\mathrm{s}}^\mathrm{max} - \lambda_{\mathrm{s}}^\mathrm{init}),
    \label{eq:GRL1}
\end{equation}
\begin{equation}
    \lambda_{\mathrm{p}}(p) = \lambda_{\mathrm{p}}^\mathrm{init} + p(\lambda_{\mathrm{p}}^\mathrm{max} - \lambda_{\mathrm{p}}^\mathrm{init}),
    \label{eq:GRL2}
\end{equation}
where $p = e_\mathrm{current}/e_\mathrm{total} \in [0,1]$ represents the relative training progress, with $e_\mathrm{current}$ and $e_\mathrm{total}$ denoting the current epoch and total number of epochs, respectively. For the pattern recognition branch, we set $\lambda_{\mathrm{s}}^\mathrm{init}=0$ and $\lambda_{\mathrm{s}}^\mathrm{max}=0.1$, while for the individual identification branch, $\lambda_{\mathrm{p}}^\mathrm{init}=1.0$ and $\lambda_{\mathrm{p}}^\mathrm{max}=1.5$.

The network was optimized using stochastic gradient descent (SGD) with separate optimizers for the two branches, using a learning rate of 0.001 and a batch size of 256.
All experiments were implemented in PyTorch 2.4.1 and conducted on a NVIDIA RTX 5000 GPU.

\subsection{Baselines}

To validate the effectiveness of the proposed method, we implemented three baseline approaches:
\begin{itemize}
    \item \textbf{Empirical risk minimization (ERM)}: A model using only the pattern recognition branch trained on the training data without adversarial components. This serves as a fundamental baseline to evaluate the contribution of our feature disentanglement strategy.
    \item \textbf{Single branch adversarial neural network (P-Only)}: A model implementing the pattern recognition branch with adversarial training, following the standard design of domain adversarial neural network (DANN)~\cite{ganin2016domain}. This baseline helps evaluate the effectiveness of our dual-branch architecture compared to conventional adversarial approaches.
    \item \textbf{Muti-task learning neural network (MTL)}: A model incorporating both pattern recognition and individual identification branches but excluding their adversarial components. This baseline is designed to assess the impact of adversarial training in our dual-branch architecture.
\end{itemize}
All baseline models were directly applied to new users without any calibration, allowing fair comparison with our proposed method.

\subsection{Performance Evaluation}
The model performance was evaluated using multiple: accuracy, area under the receiver operating characteristic curve (AUROC), and macro-F1 score (F1). These metrics were computed by averaging the results across all testing subjects for each fold. In addition, we assessed the clustering quality of learned latent features using the Davis-Bouldin index (DBI).

Let $\mathbf{z}_m^\mathrm{p}$ denote the $m$-th pattern-specific feature vector in the bottleneck layer, and $\mathbf{C}_i=\{\mathbf{z}_m^\mathrm{p}\}_{m=1}^{M_i}$ be the cluster of pattern $i$, where $M_i$ is the number of features in cluster $i$. 
The DBI is defined as
\begin{equation}
    DBI = \frac{1}{N_\mathrm{p}}\sum_{i=1}^{N_\mathrm{p}}\max\limits_{{j}\neq{i}}\left(\frac{s_i+s_j}{d_{ij}}\right),
\end{equation}
%
where $s_i$ represents the average distance of all samples in cluster $i$ to its centroid $\mathbf{c}_i$, and $d_{ij}$ denotes the distance between centroids of clusters $i$ and $j$. 
These terms are calculated as:
\begin{equation}
    s_i=\frac{1}{M_i}\sum_{m=1}^{M_i} \|\mathbf{z}_m^\mathrm{p}-\mathbf{c}_i\|,
\end{equation}
\begin{equation}
    d_{ij}=\|\mathbf{c}_i-\mathbf{c}_j\|,
\end{equation}
where the cluster centroid $\mathbf{c}_i$ is computed as:
\begin{equation}
    \mathbf{c}_i=\frac{1}{M_i}\sum_{m=1}^{M_i}\mathbf{z}_m^\mathrm{p}.
\end{equation}


\section{Results and Discussion}

In this section, we present and discuss the key results from our experiments. The performance of the proposed model is evaluated from several perspectives, including its effectiveness in EMG feature disentanglement, generalization ability on unseen user data, and the clustering capability of the learned latent features. Our findings show that the model significantly outperforms baseline methods in its ability to generalize across new individuals without requiring calibration.

\begin{figure*}[t]
    \centering
    \includegraphics[width=1.0\textwidth]{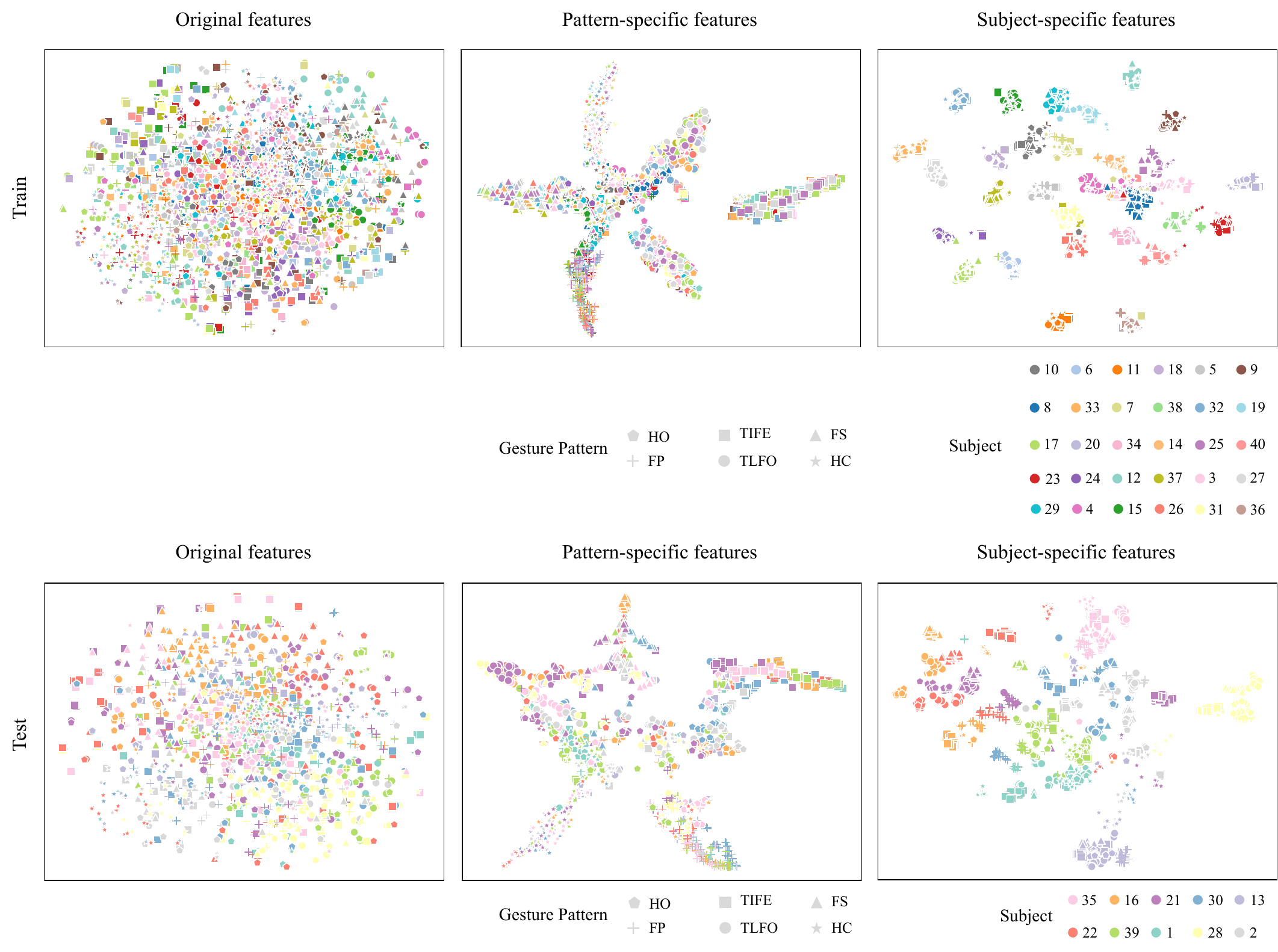} 
    \caption{$t$-SNE visualization of feature spaces learned by the proposed network. The figure shows (left) original features extracted by the shared encoder, (middle) pattern-specific features, and (right) subject-specific features for both training and test sets from the first cross-validation fold. Shapes indicate different gesture patterns while colors represent different subjects.} 
    \label{fig:tsne} 
\end{figure*}

\begin{table*}[t]
\centering
\caption{Cross-subject pattern recognition performance comparison using 4-fold validation}
\label{tab:classifcation performance}
\begin{tabular}{lcccccccccccc}
\toprule
& \multicolumn{3}{c}{Fold 1} & \multicolumn{3}{c}{Fold 2} & \multicolumn{3}{c}{Fold 3} & \multicolumn{3}{c}{Fold 4} \\
\cmidrule(lr){2-4} \cmidrule(lr){5-7} \cmidrule(lr){8-10} \cmidrule(lr){11-13}
Methods & Accuracy & F1 & AUROC & Accuracy & F1 & AUROC & Accuracy & F1 & AUROC & Accuracy & F1 & AUROC \\
\midrule
ERM     & 0.7990 & 0.9646 & 0.7855 & 0.7100 & \textbf{0.9440} & 0.6712 & 0.7336 & 0.9535 & 0.7158 & 0.7603 & \textbf{0.9593} & 0.7501 \\
P-Only  & 0.7999 & 0.9649 & 0.7831 & 0.7330 & 0.9436 & 0.7083 & 0.7556 & 0.9555 & 0.7363 & 0.7597 & 0.9539 & 0.7488 \\
MTL     & 0.7961 & 0.9626 & 0.7816 & 0.7368 & 0.9392 & 0.7122 & 0.7458 & 0.9478 & 0.7296 & 0.7483 & 0.9484 & 0.7376 \\
Proposed & \textbf{0.8201} & \textbf{0.9721} & \textbf{0.8052} & \textbf{0.7420} & 0.9416 & \textbf{0.7160} & \textbf{0.7761} & \textbf{0.9600} & \textbf{0.7597} & \textbf{0.7851} & 0.9592 & \textbf{0.7763} \\
\bottomrule
\end{tabular}
\end{table*}

\subsection{Feature Disentanglement}
To validate the model's effectiveness in EMG feature disentanglement, we employed $t$-distributed stochastic neighbor embedding ($t$-SNE) to visualize the original features extracted by the feature extractor, as well as the pattern-specific and subject-specific features extracted by the encoders $E_\mathrm{p}$ and $E_\mathrm{s}$. Different shapes were used to represent different patterns, and distinct colors were applied to differentiate subjects. Given the similarity of results across 4 folds, we present the result of Fold 1 as an example for clarity. 

As shown in Fig.~\ref{fig:tsne}, the original features are scattered and disorganized in the latent space, making it difficult to establish clear decision boundaries for either gesture recognition or individual identification. However, after processing through the following two parallel branches, the pattern-specific and subject-specific components are successfully disentangled from the original features, forming distinct clusters as shown in the $t$-SNE visualizations in their respective latent spaces. Notably, while gestures are well-separated in the pattern-specific space, some inter-cluster structure associated with subjects remains in the testing data. Similarly, in the subject-specific latent space, an inter-cluster structure corresponding to patterns is apparent, particularly in the testing data.

\subsection{Classification Performance}
The primary goal is to achieve accurate and robust EMG pattern recognition for unseen user data through feature disentanglement. Therefore, after validating the model's effectiveness in EMG feature disentanglement, it is crucial to evaluate the classification performance of the proposed model.

The quantitative results in Table~\ref{tab:classifcation performance} show that our method outperformed the baselines in the vast majority of cases, with only slight underperformance compared to ERM in terms of AUROC in Fold 2 and Fold 4. However, these differences were not statistically significant. Nevertheless, the differences among the baseline models themselves are not statistically significant. 

It is evident that ERM serves as a strong baseline, which is difficult to surpass. When employing simpler adversarial strategies or multi-task learning, improvements are not guaranteed, and in some cases, as seen with Fold 4, performance even decreases. This may be due to conflicts between the primary and adversarial branches or interference from multiple tasks.

However, the proposed method combines adversarial training and multi-task learning, guiding the network to learn towards decoupling pattern-specific and subject-specific components from the original features. In this setup, the two branches tend to interact in a mutually reinforcing manner: the good performance of the individual identification branch helps transfer subject-specific information to the pattern recognition branch in a more specific and powerful way, and vice versa. This positive interaction enhances the effectiveness of adversarial training compared to a single-branch structure. At the same time, because the learning goals of the multiple tasks become clearer through adversarial training, it helps reduce the noise in the extracted features and alleviates conflicts between the two task branches. This, in turn, enhances the overall performance of the proposed method.

\subsection{Feature Space Analysis}
\begin{figure}[t]
    \centering
    \includegraphics[width=0.49\textwidth]{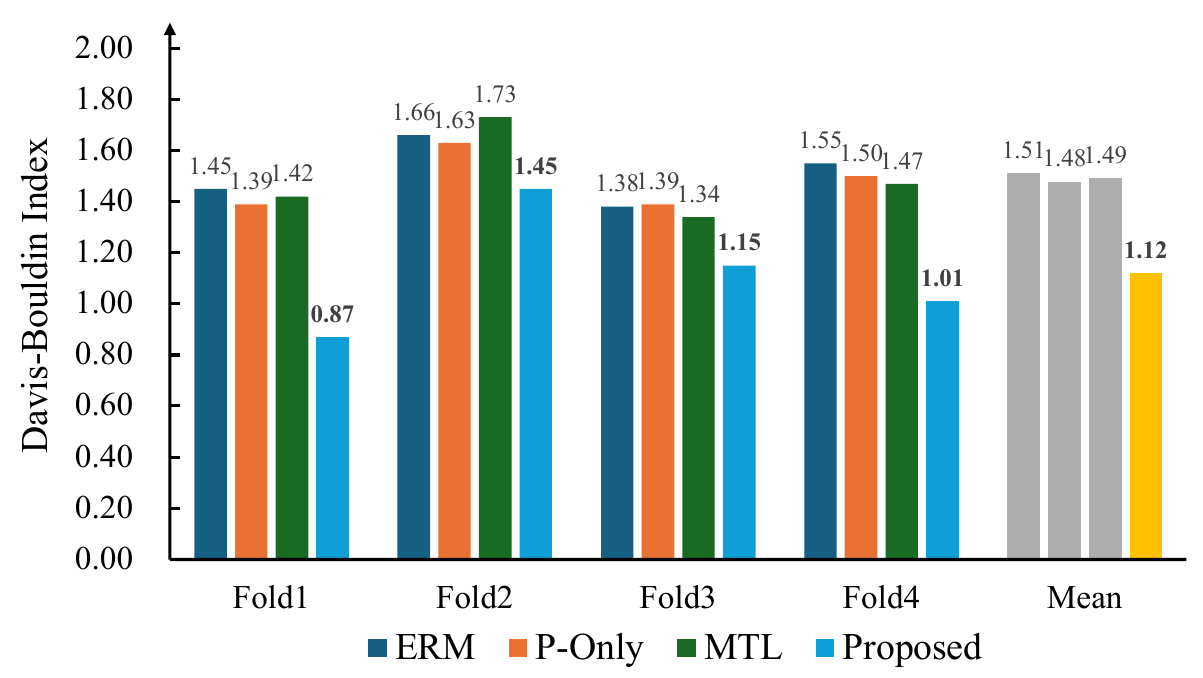} 
    \caption{Comparison of Davis-Bouldin index (DBI) across different cross-validation folds. Lower DBI values indicate better clustering quality of the extracted features.} 
    \label{fig:dbi} 
\end{figure}

To further investigate the mechanism underlying the performance improvement, we quantitatively analyzed the feature space of both the baselines and the proposed method. The assessment uses the DBI to determine whether the improvement is due to the enhanced ability of feature representation. 
As shown in Fig.~\ref{fig:dbi}, the decomposed pattern-specific features achieve an average DBI of 1.12 on the testing data, which is lower than all baselines. This result indicates that the proposed model exhibits higher intra-cluster compactness and better inter-cluster separation, reflecting its superior ability to extract both robust and discriminative features. These findings quantitatively confirm the earlier discussions that the model's performance is enhanced by a strengthened feature representation capacity through the mutually reinforcing interaction between the pattern recognition and individual identification branches.

\section{Conclusion}
In this paper, we proposed an end-to-end dual-branch adversarial neural network, in which the EMG feature extraction, feature disentanglement, pattern recognition and individual identification can be completed simultaneously. The effectiveness of EMG feature disentanglement on multi-channel EMG signals has been demonstrated, achieving cross-subject EMG pattern recognition without model calibration. This work provides new insights into cross-subject EMG pattern recognition.

Comprehensive experimental results show that our method effectively disentangles the pattern-specific and subject-specific components from the original features. Additionally, it outperforms the baseline methods across multiple classification metrics. Through analysis of the feature space, we believe that this improvement may stem from the enhanced feature representation capability.

Despite the promising results, several directions remain for enhancing the proposed method. Future work will focus on improving accuracy, validating the method’s robustness across multiple sessions and datasets, and exploring its adaptability to other downstream tasks, such as task-invariant individual identification. These efforts aim to enhance the model’s generalization capability and broaden its practical applications.

\addtolength{\textheight}{-12cm}   





\bibliographystyle{IEEEtran}
\bibliography{Ref_EMBC2025}

\end{document}